# Spatiotemporal Fusion in Remote Sensing

Hessah Albanwan[1], Rongjun Qin [1,2] *

1. Department of Civil, Environmental and Geodetic Engineering, The Ohio State University, United States, 2036 Neil Avenue, Columbus, Ohio, USA. (Albanwan.1, qin.324)@osu.edu

2. Department of Electrical and Computer Engineering, The Ohio State University, United States. qin.324@osu.edu

* Corresponding author

## Abstract

Remote sensing images and techniques are powerful tools to investigate earth surface. Data quality is the key to enhance remote sensing applications and obtaining a clear and noise-free set of data is very difficult in most situations due to the varying acquisition (e.g., atmosphere and season), sensor, and platform (e.g., satellite angles and sensor characteristics) conditions. With the increasing development of satellites, nowadays Terabytes of remote sensing images can be acquired every day. Therefore, information and data fusion can be particularly important in the remote sensing community. The fusion integrates data from various sources acquired asynchronously for information extraction, analysis, and quality improvement. In this chapter, we aim to discuss the theory of spatiotemporal fusion by investigating previous works, in addition to describing the basic concepts and some of its applications by summarizing our prior and ongoing works.

## 1. Introduction

### 1.1. Background

Obtaining a high-quality satellite image with a complete representation of earth's surface is crucial to get clear interpretability of data, which can be used for monitoring and managing natural and urban resources. However, because of the internal and external influences of the imaging system and its surrounding environment, the quality of remote sensing data is often insufficient. The internal imaging system conditions such as the spectral characteristics, resolution, and other factors of the sensor, algorithms used to calibrate the images, etc. The surrounding environment refers to all external/environmental influences such as weather and season. These influences can cause errors and outliers within the images, for instance, shadow and cloud may cause obstructions in the scene and may occlude part of the information regarding an object. These errors must be resolved in order to produce high-quality remote sensing products (e.g., land-cover maps).

With the rapid and increasing development of satellite sensors and their capabilities, studies have shown that fusion of images from multi-source, multi-temporal, or both is the key to recover the quality of a satellite image. Image fusion is known as the task of integrating two or more images into a single image [2,3]. The fusion of data essentially utilizes redundant information from multiple images to resolve or minimize uncertainties associated with the data, with the goals to reject outliers, to replace and fill missing data points, and enhance spatial and radiometric resolutions of the data, etc. Fusion has been used in a wide range of remote sensing applications such as radiometric normalization, classification, change detection, etc. In general, there are two types of fusion algorithms, being spatial-spectral [4–7] and spatiotemporal fusion [8–10]. The spatial-spectral fusion uses the local information in a single image to predict the pixels true values based on spectrally similar neighboring pixels. It is used for various types of tasks and applications such as filling missing data (also known as image inpainting) and generating high-resolution images (e.g. Pan-sharpening [11] and super-resolution [12]). It can include filtering approaches such as fusing information within a local window using methods such as interpolation [13, 14], maximum a posteriori (MAP), Bayesian model, Markov random fields (MRFs), and Neural Networks (NN) [4, 12, 15–18]. Although spatial-spectral fusion is efficient, it is not able to incorporate information from temporal images which produce dramatic radiometric differences such as those introduced by meteorological, phenological, or ecological changes. For instance, radiometric distortions and impurities in an image due to metrological changes (e.g., heavy cloud cover, haze, or shadow) cannot be entirely detected and suppressed by the spatial-spectral fusion since it only operates locally within a single image. To address this issue, researchers suggested spatiotemporal fusion which encompasses the spatial-spectral fusion and offers a filtering algorithm that is invariant to dynamic changes over time, in addition to being robust against noise and radiometric variations. Identifying spatiotemporal patterns is the core to spatiotemporal fusion, where the patterns are intended to define a correlation between shape, size, texture, and intensity of adjacent pixels across images taken at different time, of different types, and from different sources.

   Spatiotemporal fusion has been an active area of study over the last few decades [9]. Many studies have shown that maximizing the amount of information through integrating the spatial, spectral, and temporal attributes can lead to accurate stable predictions and enhance the final output [8, 9, 19–21]. Spatiotemporal fusion can be applied within local and global fusion frameworks, where locally it can be performed using weighted functions and local windows around all pixels [22–24], and globally using optimization approaches [25, 26]. Additionally, spatiotemporal fusion can be performed on various data processing levels depending on the desired techniques and applications to be used [3]. It also can depend on the type of data used, for instance, per-pixel operations are well suited for images acquired from the same imaging system (i.e. same sensor) since they undergo a similar calibration process and minimum spectral differences in terms of having the same number of bands and bandwidths ranges in the spectrum, whereas feature or decision level fusion are more flexible and able to handle heterogeneous data such as combing elevation data (e.g. LiDAR) with satellite images [27]. Fusion levels include:

***Pixel-level image fusion*** is a direct low-level fusion approach. It involves pixel-to-pixel operation, where the physical information (e.g. intensity values, elevation, thermal values, etc.) associated with each pixel within two or more images is integrated into a single value [2]. It includes methods such as spatial and temporal adaptive reflectance fusion model (STARFM), Spatial and Temporal Reflectance Unmixing Model (STRUM), etc. [22–24].

***Feature-level image fusion*** involves extracting and matching distinctive features from two or more overlapping images using methods such as dimensionality reduction like principal component analysis (PCA) and Linear Discriminant Analysis (LDA), SIFT, SURF, etc. [2, 28]. Fusion is then performed using the extracted features and the coefficients corresponding to them [2, 29]. Some other common methods that include spatiotemporal fusion on feature-level are sparse representation and deep learning algorithms [10, 30], [31], [32–38].

***Decision-level image fusion*** is a high-level fusion that requires each image to be processed individually until an output (e.g., classification map). The outputs are then post-processed using decision-level fusion techniques [2, 39]. This level of fusion can include the previous two levels of fusion (i.e. per-pixel operations or extracted features) within its operation [40, 41].

In this chapter, we will focus on the concept of the spatiotemporal-based fusion method and highlight its applications on all levels of fusion. We will discuss all aspects of spatiotemporal fusion starting from its concepts, preprocessing steps, the approaches, and techniques involved. We will also discuss some examples that apply spatiotemporal fusion for remote sensing applications.

### 1.2. Contributions

This book chapter introduces the spatiotemporal analysis in the fusion algorithms to improve the quality of remote sensing images. We will explore spatiotemporal fusion advantages and limitations, as well as their applications and associated technicalities under three scenarios:

1) Pixel-level spatiotemporal fusion
2) Feature-level spatiotemporal fusion
3) Decision-level spatiotemporal fusion

### 1.3. Organization

The organization of this chapter is as follows: Section 2 describes remote sensing data and acquisition and generation processes and necessary preprocessing steps for all fusion levels. Section 3 talks about spatiotemporal fusion techniques under the three levels of fusion: pixel-level, feature-level, and decision-level, which can be applied to either multi-source, multitemporal, or multi-source multitemporal satellite images. Section 4 describes some applications applying spatiotemporal fusion, and finally, Section 5 concludes the chapter.

## 2. Generic steps to spatiotemporal fusion

The spatiotemporal analysis allows the investigation of data from various times and sources. The general workflow to any spatiotemporal fusion process is shown in Figure 1. The process description towards a fused image is demonstrated in Figure 1(a), where it starts with the input acquisition, preprocessing, and finally fusion. Data in remote sensing is either acquired directly from a sensor (e.g. satellite images) or indirectly generated using algorithms (e.g. depth image from dense image matching algorithms [42]) (see Figure 1(b)). It also includes data from single or multiple sources (see Figure 1(b)), however, combing multi-source and multi-temporal images requires preprocessing steps to assure data consistency for analyses. The preprocessing steps can include radiometric and geometric correction and alignment (see Figure 1(a)). The main spatiotemporal fusion algorithm is then performed using one or more of the three levels of fusion as a base for their method. In this section, we will discuss the most common preprocessing steps to spatiotemporal fusion, as well as the importance and previous techniques used in spatiotemporal fusion in the three levels of fusion to improve the quality of images and remote sensing applications.

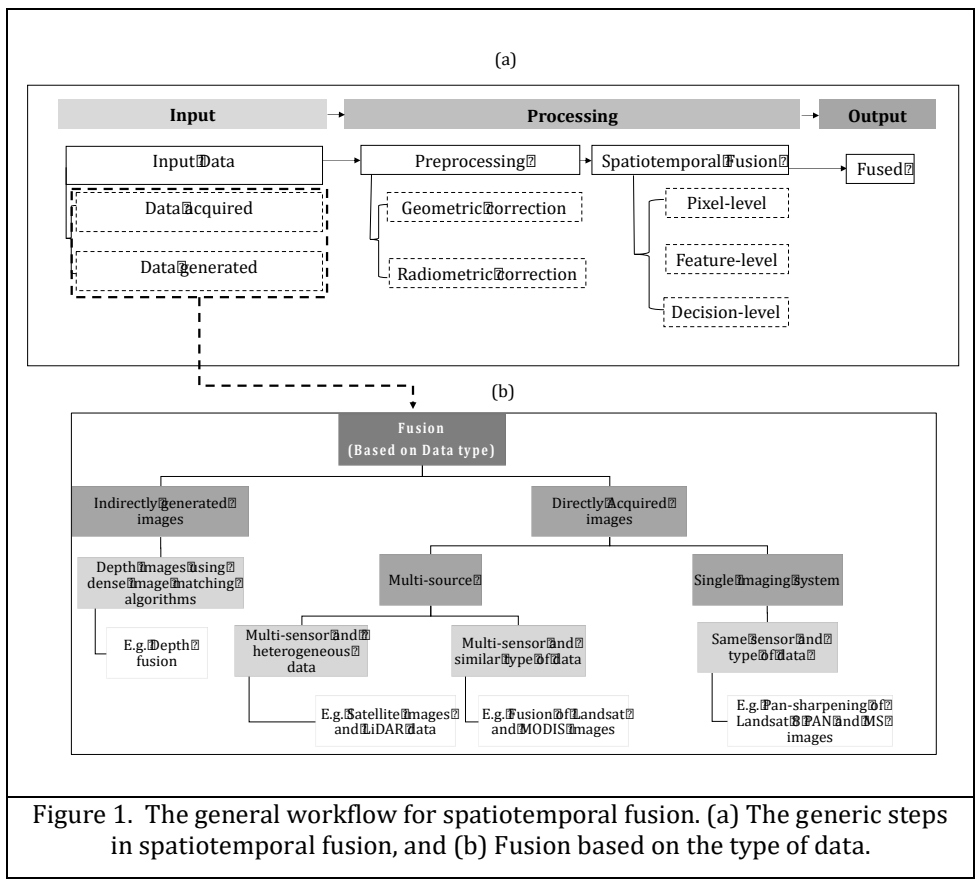

Figure 1. The general workflow for spatiotemporal fusion. (a) The generic steps in spatiotemporal fusion, and (b) Fusion based on the type of data.

### 2.1. Data acquisition and generation

Today, there exists a tremendous number of satellite sensors with varying properties and configurations providing researchers with access to a large amount of satellite data. Remote sensing images can be acquired directly from sensors or indirectly using algorithms. It is also available with a wide range of properties and resolutions (i.e., spatial, spectral, and temporal resolutions), which are described in detail in Table1.

| Type of resolution | Spatial resolution | Spectral resolution | Temporal resolution |
|---|---|---|---|
| Definition | Describes the ground area covered by a single pixel in the satellite images. It is also known as the ground sampling distance (GSD) and can range from a few hundred meters to sub-meters. Satellite sensors like Moderate Resolution Imaging Spectroradiometer (MODIS) produce coarse resolution images with 250-, 500-, and 1000-meters, while fine resolution images are produced by satellite like very high-resolution (VHR) with sub-meter level[43]. | Refers to the ability of satellite sensors to capture images with wide ranges of the spectrum. It includes hyperspectral (HS) images with thousands of bands or multispectral (MS) images with few numbers of bands (up to 10-15 bands) [43]. It may also include task-specific bands that are beneficial to study the environment and weather like the thermal band as in Landsat 7 thematic mapper plus (ETM+) [43]. The spectral resolution also refers to the wavelength interval over in the spectral signal domain, for instance, MODIS has 36 bands falling between 0.4 to 14.4 $\mu m$, whereas Landsat 7 (ETM+) has 7 bands ranging from 0.45 to 0.9 $\mu m$. | Is the ability for satellite sensors to capture an object or phenomena on certain periods of time, also known as the revisiting time of sensor to a certain location on the ground. Today, modern satellite systems allow monitoring earth surface over short and regular periods of time, for instance, MODIS provides almost a daily, while Landsat covers the entire earth surface every 16 days. |

Table 1. Satellite sensors characteristics and resolutions.

### 2.1.1. Data acquisition

Generally, there exist two types of remote sensing sensor systems, being active and passive sensors [43]: ***Active sensors*** record the signal that is emitted from the sensor itself and received back when it reflects off the surface of the earth. It includes sensors like Light Detection and Ranging (LiDAR) and Radar. ***Passive sensors*** record the reflected signal off the ground after being emitted from a natural light source like the Sun. It includes satellite sensors that produce satellite images such as Landsat, Satellite Pour l'Observation de la Terre (SPOT), MODIS, etc.

### 2.1.2. Data generation

Sometimes in remote sensing, the derived data can be also taken as measurements. For example, depth images with elevation data are derived through photogrammetric techniques on satellite stereo or multi-stereo images [42], classification maps, change detection maps, etc. In this section, we will discuss two important examples of the commonly fused remote sensing data and their generation algorithms:

**1) Depth maps (or Digital Surface Model (DSM))**
3D geometric elevation information can either be obtained directly using LiDAR or indirectly using dense image matching algorithms such as Multiview stereo (MVS). However, because LiDAR data is expensive and often unavailable for historic data (before 1970s when LiDAR was developed), generating depth images using MVS algorithms is more convenient and efficient. MVS algorithms include several steps:

***Images acquisition and selection*** to perform MVS algorithm requires having at least a pair or more of overlapping images that are captured from different viewing angles to assure selecting an adequate number of matching features. Specifically, this refers to the process of feature extraction and matching, where unique features are being detected and matched in pairs of images using feature detectors and descriptors methods such as Harris, SIFT, or SURF [44].

***Dense image matching and depth map generation***. Dense image matching refers to the processing of producing dense correspondences between two or among multiple images, and with their pre-calculated geometrical relationship, depth/height information can be determined through ray triangulation [45]. The dense correspondences problem, with pre-calculated image geometry, turns to a 1-D problem in rectified image (also called epipolar image) [46], called disparity computation, which is basically the difference between the left and right views as shown below:

$$Disparity = \Delta x = x_l - x_r = \frac{f T}{z} \quad (1)$$

Where $x_l$ and $x_r$ are distance of pixel in the left and right images accordingly, f is the focal length, $T$ is the distance between the cameras, and $z$ is the depth. The depth ($z$) is then estimated from equation [1] by taking the focal length times the distance between the cameras divided by the disparity as follows:

$$Depth = Z = \frac{ft}{|x_l - x_r|} \quad (2)$$

In addition, it is noted that assessing and selecting good pairs of images can improve dense image matching and produce more accurate and complete 3D depth maps [47, 48].

**2) <u>Classification maps</u>**
Image classification can be divided into two categories: **1) Supervised classification** is a user-guided process, where classification depends on prior knowledge about the data that is extracted from the predefined training samples by the user, some popular supervised classification methods include support vector machine (SVM), random forest (RF), decision trees DT, etc. [49–51]. 2) **Unsupervised classification** is a machine-guided process, where the algorithms classify the pixels in the image by groups of similar pixels to come up with specific patterns that define each class. These techniques include segmentation, clustering, nearest neighbor classification, etc. [49].

## 2.2. Preprocessing steps

### 2.2.1. Geometric correction

Image registration and alignment is an essential preprocessing step to any remote sensing application that processes two or more images. For accurate analyses of multi-source multi-temporal images, it is necessary that overlapping pixels in the images correspond to the same coordinates or points on the earth's surface. Registration can be performed manually by selecting control points (CP) between a pair of images to determine the transformation parameters and wrap the images with respect to a reference image [52]. An alternative approach is an automated CP extraction that operates based on mutual information (MI) and similarity measures of the intensity values [52]. According to [53], there are a few common and sequential steps for image registration including the following steps:

***Unique feature selection, extraction, and matching*** refers to the process where unique features are detected using feature extraction methods, then matched to their correspondences in a reference image. A feature can be a shape, texture, intensity of a pixel, edge, or index such as vegetation and morphological index. According to [54] and [55], features can be extracted based on the content of a pixel (e.g. intensity, depth value, or even texture) using methods such as SIFT, difference of Gaussian (DOG), Harris detection, Histogram of oriented gradient (HOG), etc. [53, 56–58] or based on a patch of pixels [59–61] like using deep learning methods (e.g. convolutional neural networks (CNNs)), which can be used to extract complete objects to be used as features.

***Transformation*** refers to the process of computing the transformation parameters (e.g., rotation, translation, scaling, etc.) necessary to convolve an image to a coordinate system that matches a reference image. The projection and transformation methods include similarity, affine, projective, etc. [53].

***Resampling*** is the process where an image is converted into the same coordinate system as the reference image using the transformation parameters; it includes methods such as interpolation, bilinear, polynomial, etc. [53].

### 2.2.2. Radiometric correction

Radiometric correction is essential to remove spectral distortion and radiometric inconsistencies between the images. It can be performed either using absolute radiometric normalization (ARN) or relative radiometric normalization (RRN) [62–64]. The ARN requires prior knowledge of physical information related to the scene (e.g. weather conditions) for normalization [63, 65–67]. While, RRN radiometrically normalize the images based on a reference image using methods such as dark object subtraction (DOS), histogram matching (HM), simple regression (SR), pseudo-invariant features (PIF), iteratively re-weighted MAD transformation, etc. [62, 64, 68].

## 3. Data analysis and spatiotemporal fusion

Pixels in remote sensing data are highly correlated over space and time due to earth surface characteristics, repeated patterns (i.e., close pixels belong to the same object/class), and dynamics (i.e., season). The general algorithm for spatiotemporal fusion is demonstrated in Figure 2, where all levels of fusion follow the same ideology. The minimum image requirement for spatiotemporal fusion is a pair of images whether they are acquired from multiple sources or time, the input images are represented with $t_1$ to $t_n$ in Figure2. The red square can be either a single raw pixel, an extracted feature vector, or a processed pixel with valuable information (e.g., probability value indicating the class of a pixel). The fusion algorithm then finds the spatiotemporal patterns over space (i.e., the coordinates ($x, y$), and pixel content) and time ($t$) to predict accurate and precise values of the new pixels (see Figure 2). In this section, we will provide an overview of some previous works regarding spatiotemporal image fusion that emphasize the importance of space-time correlation to enhance image quality and discuss this type of fusion under the context of three levels of fusion: pixel-level, feature-level, and decision-level.

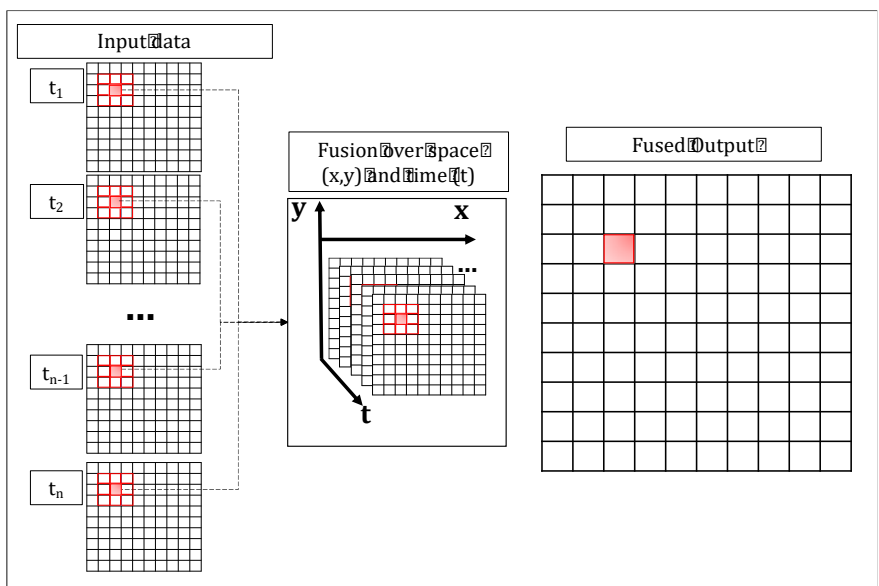

Figure 2. The general concept of spatiotemporal fusion to process a patch of pixels (the red square) spatially across different times ($t$).

### 3.1. Pixel-level spatiotemporal fusion

As mentioned in the introduction, pixel-based fusion is the most basic and direct approach to fuse multiple images by performing pixel-to-pixel operations; it has been used in a wide range of applications and is preferred because of its simplicity.

Many studies performing pixel-level fusion algorithms realized the power of spatiotemporal analysis in fusion and used it in a wide range of applications such as monitoring, assessing, and managing natural sources (e.g. vegetation, cropland, forests, flood, etc.), as well as, urban areas [9]. Most of the pixel-level spatiotemporal fusion algorithms operate as filtering or weighted-function method, they process a group of pixels in a window surrounding each pixel to compute the corresponding spatial, spectral, and temporal weights (see Figure 3). A very popular spatiotemporal fusion method that sets the base for many other fusion methods is spatial and temporal adaptive reflectance fusion model (STARFM); it is intended to generate a high-resolution image with precise spectral reflectance by merging multi-source fine and coarse resolution images [22]. Their method resamples the coarse resolution MODIS image to have a matching resolution as the Landsat TM image, after that it computes the overall weight by calculating the spectral and temporal differences between the images. STARFM is highly effective for detecting phenological changes, but it fails to handle heterogeneous landscapes with rapid land cover changes and around mixed pixels [22]. To address this issue, [20] have proposed Enhanced STARFM (ESTARFM) that applies a conversion coefficient to assess the temporal differences between fine and coarse resolution images. In [69], Hilker also addressed the problem of drastic land cover change by proposing Spatial Temporal Adaptive Algorithm for mapping Reflectance Change (STAARCH) that applies Tasseled cap transformation [70] to detect the seasonal changes over a landscape. For further improvement of these algorithms, studies have suggested using machine learning methods to identify similar pixels by their classes [71]. They also show an example of using machine learning unsupervised classification within the spatiotemporal fusion to enhance its performance. They used clustering on one of the images using ISODATA method [72], where pixels are considered similar if the difference between the current and central pixel in the window is less than one standard deviation of the pixels in the cluster. Other methods use filtering algorithms to enhance the spatial and spectral aspects of images, in addition to embedding the temporal analysis to further enhance the quality and performance of an application. For instance, [73] proposed a method that combines the basic bilateral filter with STARFM to estimate Land surface temperature (LST). In [19], they proposed a 3D spatiotemporal filtering as a preprocessing step for relative radiometric normalization (RRN) to enhance the consistency of temporal images. Their idea revolves around finding the spatial and spectral similarities using a bilateral filter, followed by assessing the temporal similarities for each pixel against the entire set of images. The temporal weight which assesses the degree of similarity is computed using an average Euclidean distance using the multitemporal data. In addition to the weighted-based functions, approaches such as unmixing-based and hybrid-based methods are also common in spatiotemporal fusion [74]. The unmixing-based methods predict the fine resolution image reflectance by computing the mixed pixels from coarse resolution image [75], while hybrid-based methods use a color mapping function that computes the transformation matrix from the coarse resolution image and applies it on the finer resolution image [76].

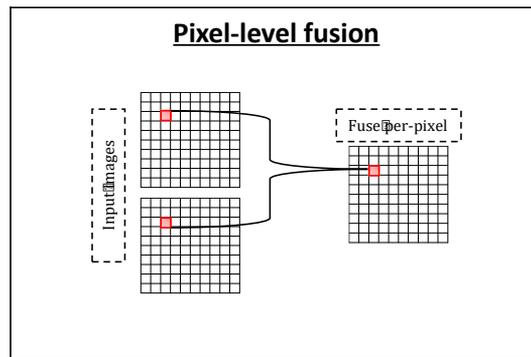

Figure 3. Pixel-based fusion process diagram.

## 3.2. Feature-level spatiotemporal fusion

Feature-level fusion is a more complex level of fusion, unlike pixel-based operations, it can efficiently handle heterogeneous data that vary in modality and source. According to [2], feature-based fusion can either be conducted directly using semantically equivalent features (e.g., edges) or through probability maps that transform images into semantically equivalent features. This characteristic allows fusion to be performed regardless of the type and source of information [27]. Fusion can then be performed using arithmetic (e.g., addition, division, etc.) and statistical (e.g., mean, median, maximum, etc.) operations; the general process of feature-based fusion is shown in Figure 4. The approach in [27] demonstrates a simple example on feature-level spatiotemporal fusion to investigate and monitor deforestation; in their method, they combined data from medium resolution synthetic aperture radar (SAR) and MS Landsat data, they extracted features related to vegetation and soil location (using scattering information and Normalized Difference Fraction Index (NDFI) respectively), finally, fusion was performed through decision tree classifier. Both [26] and [62] point out to the most popular methods in feature-level fusion which include the Laplacian pyramid, gradient pyramid, and morphological pyramid, high-pass filter, and wavelet transform methods [77–81]. A very famous fusion example in this category is inverse discrete wavelet (IDW) transform, which is a wavelet transform fusion approach; it uses temporal images with varying spatial resolutions to down-sample the coarse resolution image. It basically extracts the wavelet coefficients from the fine resolution image and uses them to down-sample the coarse resolution image [82]. Sparse representation is another widely used learning-based method in feature-level fusion due to its good performance [8, 10, 30, 31]. All sparse representation algorithms share the same concept and core idea, where the general steps include: 1) divide the input images into patches, 2) extract distinctive features from the patches (e.g. high-frequency feature patches), 3) generate coefficients from the feature patches, 4) training jointly using dictionaries to find similar structures by extracting and matching feature patches, finally, 5) fusion using the training information and extracted coefficients [8, 10, 30, 79, 83, 84].

Another state-of-the-art approach in feature- and decision-level fusion is deep learning or artificial neural networks (ANN). They are currently a very active area of interest in many remote sensing fields (especially image classification) due to their outstanding performance that surpasses traditional methods [32–38,76, 82,83, 84]. They are also capable of dealing with multi-modality like images from varying sources and heterogeneous data, for instance, super-resolution and pan-sharpening images from different sensors, combining HS and MS images, combining images with SAR or LiDAR data, etc. [32–38]. In feature-level fusion, the ANN is either performed on the images for feature extraction or to learn from the data itself [38]. The extracted features from the temporal images or classification map are used as an input layer, which are then weighted and convoluted within several intermediate hidden layers to result in the final fused image [32–35, 37, 79]. For instance, [85] uses neural networks (CNN) to extract features from RGB image and a DSM elevation map, which are then fed into the SVM training model to generate an enhanced semantic labeling map. ANNs have also been widely used to solve problems related to change detection of bi-temporal images such as comparing multi-resolution images [86] or multi-source images [87], which can be solved in a feature-learning representation fashion. For instance, the method in [87] directly compares stacked features extracted from a registered pair of images using deep belief networks (DBN)

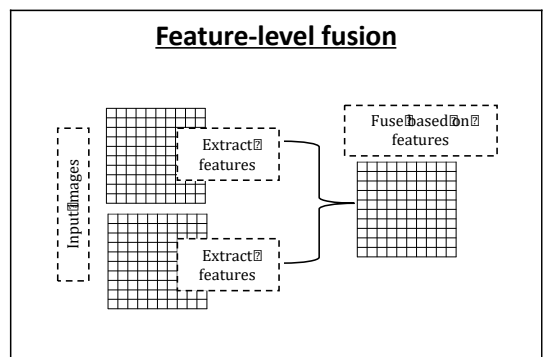

Figure 4. Feature-based fusion diagram.

## 3.3. Decision-level spatiotemporal fusion

The decision-level fusion operates on a product-level, where it requires images to be fully and independently processed until the meaningful output (e.g., classification or change detection maps) (see Figure 5). Decision-level fusion can adapt to different modularities like combing heterogeneous data such as satellite and depth images, which can be processed to common outputs (e.g. full/partial classification maps) for fusion [88]. Additionally, the techniques followed by this fusion type are often performed under the umbrella of Boolean or statistical operations using methods like likelihood estimation, voting (e.g. majority voting, Dempster-Shafer's estimation, fuzzy Logic, weighted sum, etc.) [88–90]. In [88], they proc an example on the mechanism of decision-level fusion; they developed a fusion approach to detect cracks and defects on the ground surface, they first convert multitemporal images into spatial density maps using kernel density estimation (KDE), then, fused the pixels density values using a likelihood estimation method. In general, most of the decision-level fusion techniques rely on probabilistic methods, where they require generating an initial label map with each pixel upholding a probability value and indicating its belonging to a certain class, which can be generated using traditional classification methods like the supervised (e.g., random forest) or unsupervised (e.g., clustering or segmentation) classification (see Section 2.1.2.). Another advantage of the decision-level fusion is that it can be implemented while incorporating both levels of fusion the pixel- and feature-level. The method in [41], shows a spatiotemporal fusion algorithm that includes all levels of fusion, where they propose a post-classification refinement algorithm to enhance the classification maps. First, they generate probability maps for all temporal images using a random forest classifier (as an initial classification map), then they use a recursive approach to iteratively process every pixel in the probability maps by fusing the multitemporal probability maps with the elevation from the DSMs using a 3D spatiotemporal filtering. Similarly, [40] have also proposed fusion of probability maps for building detection purposes, where they first generate the probability maps, then fuse them using a simple 3D bilateral filter.

Recently, more focus has been driven toward using spatiotemporal fusion to recover the quality of 3D depth images generated from MVS (e.g., DSM fusion). Median filtering is the oldest and most common fusion approach for depth images, it operates by computing the median depth of each pixel from a group of pixels at the same location in the temporal images [91]. The median filtering is robust to outliers and is efficient in filling missing depth values. However, the median filter only exploits the temporal domain, to further enhance its performance and the precision of the depth values, studies suggest spatiotemporal median filtering. In [92], they have proposed adaptive median filtering that operates based on the class of the pixels, they use an adaptive window to isolate pixels belonging to the same class, then choose the median pixel based on the location (i.e. adaptive window) and temporal images. In [93], they also show that spatiotemporal median filtering can be improved by adopting an adaptive weighing-filtering function that involves assessing the uncertainty of each class in the spatial and temporal domains in the depth images using standard deviation. The uncertainty will then be used as the bandwidth parameter to filter each class individually. The authors in [47] also suggested a per-pixel fusing technique to select the depth value for each pixel by using a recursive K-median clustering approach that generates 1-8 clusters until it reaches the desired precision.

Another complex yet efficient methods used in decision-level fusion are deep learning algorithms as mentioned previously in Section 2.3.2. [94]. They are either used as postprocessing refinement approaches or to learn end-to-end from a model [38]. For example, The method in [95] used a postprocessing enhancement step for semantic labeling, where they first generate probability maps using two different methods RF and CNN using multimodal data (i.e. images and depth images), then they fused the probability maps using Conditional random fields (CRFs) as postprocessing approach. In [96], on the other hand, used a model learning-based method, where they first semantically segment multisource data (i.e. image and depth image) using a SegNet network, then fuse their scores using a residual learning approach.

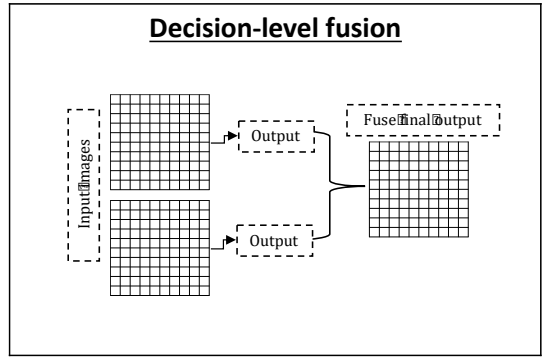

Figure 5. Feature-based fusion diagram.

## 4. Examples on spatiotemporal fusion applications

### 4.1. Spatiotemporal fusion of 2D images

*Background and objective*

A 3D spatial-temporal filtering algorithm is proposed in [19] to achieve relative radiometric normalization (RRN) by fusing information from multi-temporal images. RRN is an important preprocessing step in any remote sensing application that requires image comparison (e.g., change detection) or matching (e.g., image mosaic, 3D reconstruction, etc.). RRN is intended to enhance the radiometric consistency across a set of images, in addition to reducing radiometric distortions that result due to sensor and acquisition conditions (as mentioned in Section 1.1.). Traditional RRN methods use a single reference image to radiometrically normalize the rest of the images. The quality of the normalized images highly depends on the reference image, which requires the reference image to be noise-free or to have minimum radiometric distortions. Thus, the objective of [19] is to generate high-quality radiometrically consistent images with minimum distortion by developing an algorithm that fuses the spatial, spectral, and temporal information across a set of images.

*Theory*

The core of the 3D spatiotemporal filter is based on the bilateral filter, which is used to preserve spectral and spatial details. It is a weighting function that applies pixel-level fusion on images from multiple dates (see Figure 4. (a)). The general form of this filter is as follows:

$$\bar{I}_i = \int_\Omega w_{j,i} \cdot I_j \cdot dj \qquad (1)$$

Where the original and filtered images are indicated using $I$ and $\bar{I}$. The weight for every pixel at point j into the fused pixel $i$ is indicated using $w_{j,i}$. The filtering is carried out on the entire space of a set of images $\Omega$ including all domains the spatial (i.e., pixels' coordinates $(x, y)$), the spectral (i.e., intensity value), and temporal (i.e., intensity of temporal images). The spatial and spectral weights are described by [97] and are indicated in Eq.(3) and Eq.(4) respectively

$$w_{spatial} = exp\left(\frac{|j_x - i_x|^2}{\sigma_{\vec{x}}} + \frac{|j_y - i_y|^2}{\sigma_{\vec{x}}}\right), j, i \in \Omega \quad (3)$$

$$w_{spectral} = exp\left(-\frac{|I_j - I_i|^2}{\sigma_I}\right), j, i \in \Omega \quad (4)$$

Where, $I$ is the pixel value at x and y location, and $\sigma_{\vec{x}}$ and $\sigma_I$ are the spatial and spectral bandwidths respectively that set the degree of filtering based on the spatial and spectral similarities between the central pixel and nearby pixels. The novelty of this filter is in the design of the temporal weight, where it computes the resemblance between every image and the entire set of images using an average Euclidean distance as the following

$$w_{spectral} = exp\left(-\frac{|j_t - i_t|^2}{\sigma_T}\right), j, i \in \Omega \quad (5)$$

Where $(j_t - i_t)$ are the difference between the current image being processed and all other images and $\sigma_T$ is the degree of filtering along the temporal direction. Eq. (5) allows all images to contribute toward each other in enhancing the overall radiometric characteristics and consistency without requiring a reference image for the RRN process.

*Experimental Results and analysis*

The 3D spatial-temporal filter was conducted on three experiments with varying resolutions and complexities. Experiments 1 and 2 were applied on urban and sub-urban areas respectively each experiment had 5 medium resolution images from Landsat 8 satellite each (with 15-30 meters spatial resolution). Experiment 3 was on a fine resolution image from Planet satellite (with 3-meters spatial resolution). Figure 6(b) and (c) shows an example for the input and results of the filter using the data from experiment 1 (i.e., the urban area). The input images show a significant discrepancy in the radiometric appearance (see Figure 6(b))., however, the heterogeneity between multitemporal images is reduced after the filter (see Figure 6(c)). By comparing the original and filtered images in Figure 6(c), we can notice that the land covers are more similar in the filtered images than in the original images. For instance, the water surface (shown in Figure 6(c) in blue bold dashed line) used to have a clear contrast in intensity in the original images, but after the filtering process, they become more spectrally alike in terms of intensity looks and ranges.

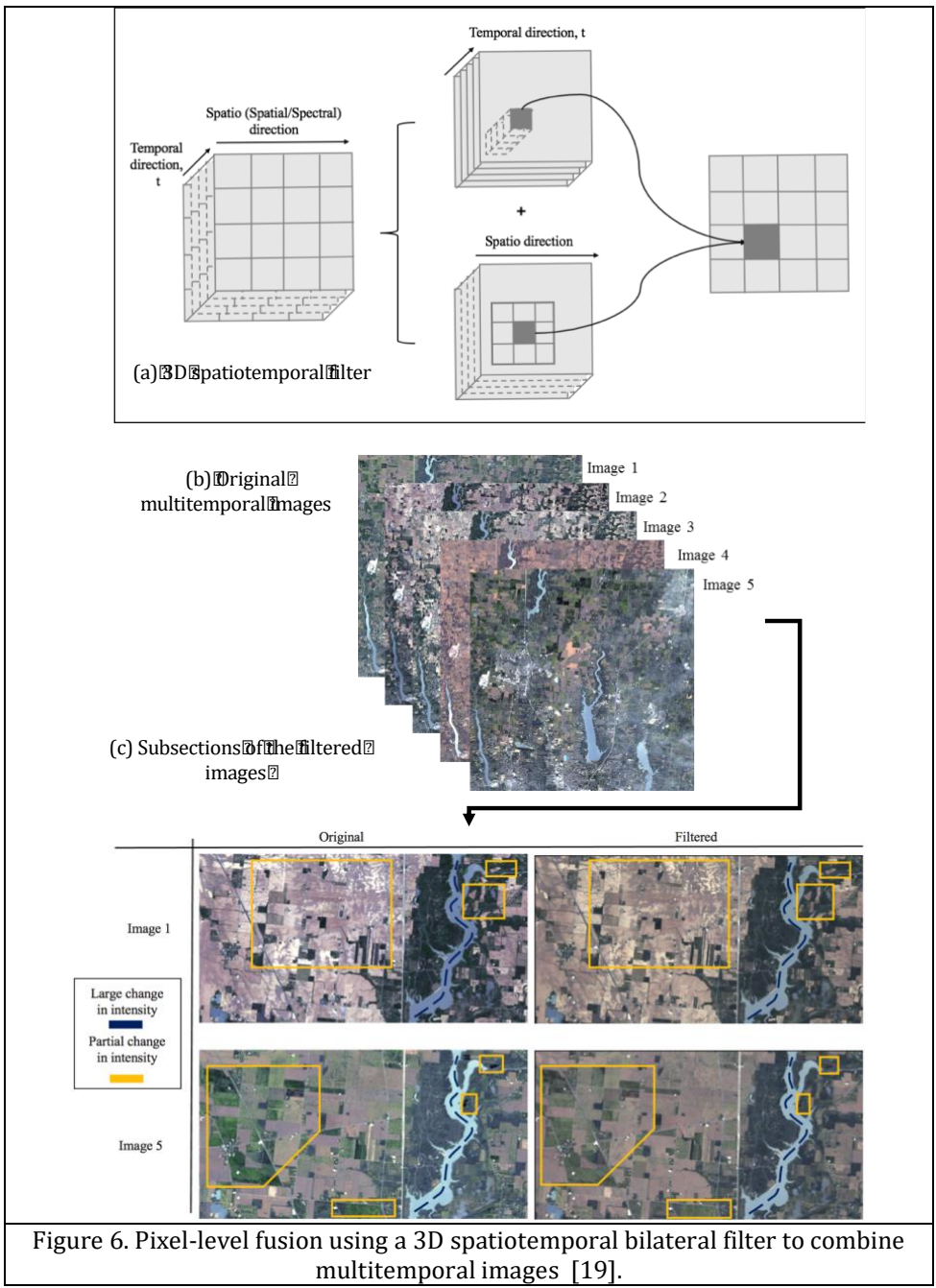

Figure 6. Pixel-level fusion using a 3D spatiotemporal bilateral filter to combine multitemporal images [19].

The experiments are also validated numerically using transfer learning classification (using SVM) to test the consistency between the normalized filtered images. The transfer learning classification uses reference training data from one image and applies it to the rest of the images. The results in Table 2. indicate that the filtered images have higher accuracy than the nonfiltered original images, where the average improvement in accuracy is ~6%, 19%, and 2 % in all three experiments, respectively. Reducing the uncertainty in the filtering process by not requiring a reference image for normalization was the key to this algorithm. The algorithm was formulated to take advantage of the temporal direction by treating all images in the dataset as a reference. Therefore, it will have higher confidence to distinguish between actual objects and radiometric distortions (like clouds) in the scene when processing each pixel.

|  | Transfer learning classification | | | | |
|---|---|---|---|---|---|
| Image | 1 | 2 | 3 | 4 | 5 |
| Exp. I Suburban | | | | | |
| Without filter | 80.88 | 74.14 | 93.30 | 93.59 | 91.97 |
| With filter | **91.99** | **91.96** | **93.95** | 86.08 | **94.30** |
| Exp. I - Urban | | | | | |
| Without filter | 72.61 | 67.91 | 81.00 | 50.21 | 93.75 |
| With filter | **89.29** | **90.60** | **91.35** | **77.82** | 93.10 |
| Exp. II | | | | | |
| Without filter | 66.48 | 74.14 | 68.35 | 67.17 | 73.30 |
| With filter | **66.75** | **76.20** | **72.06** | 65.10 | **78.54** |

Table 2. The accuracy results for the 3D spatial-temporal filter. The bold numbers indicate an increase in the accuracy, and the numbers highlighted in gray indicate the reference image used for the training in the transfer learning classification [19].

## 4.2. Spatiotemporal fusion of multi-source multi-temporal images

*Background and objective*

Multitemporal and multi-source satellite images often generate inconsistent classification maps. Noise and misclassifications are inevitable when classifying satellite images, the precision and accuracy of classification maps vary based on the radiometric quality of the images. The radiometric quality is a function of the acquisition and sensor conditions as mentioned in the background in section 1.1. The algorithm can also play a major role in the accuracy of the results, some classification algorithms are more efficient than others, while some can be sensitive to the spatial details in the images like complex dense areas and repeated patterns, which lead objects of different classes to have similar spectral reflectance. The acquisition time, type of algorithm, and distribution of objects in the scene are huge factors that can degrade the quality and generate inconsistent classification maps across different times. To address these issues, the authors in [41] proposed a 3D iterative spatiotemporal filtering to enhance the classification maps of multitemporal very high resolution satellite images. Since the 3D geometric information is more stable and is invariant to spectral changes across temporal images, [41] proposed combining the 3D geometric information in the DSM with multitemporal classification maps to provide spectrally invariant algorithm.

*Theory*

The 3D iterative spatiotemporal filter is a fusion method that combines information from various types, sources, and times. The algorithm is a combination of feature- and decision-levels of fusion; it is described in detail in Algorithm 1. The first step is to generate initial probability maps for all images using random forest classification. The inference model is then built to recursively process every pixel in the probability maps using a normalized weighting function that computes the total weight $W_{3D}(x_j, y_j, t_n)$ based on the spatial ($W_{spatial}$), spectral ($W_{spectral}$), and temporal ($W_{temproral}$) similarities. The temporal weight is based on the elevation values in the DSMs. The probability value for every pixel is computed and updated using $W_{3D}(x_j, y_j, t_n)$ and the previous iteration until it satisfies the convergence condition, which requires the difference between the current and previous iterations to be under a certain limit.

---

**Algorithm 1: Pseudocode of the proposed 3D iterative spatiotemporal filter** [41]

**Input:** Initial probability maps $P_c^0(x_i, y_i, t_n)$, orthophotos I, and the bandwidths: $\sigma_s$ and $\sigma_r$
**Output:** Final probability maps $P_c^f(x_i, y_i, t_n)$

*For* every category/class c *do*
   *While* not converge *do*
      *For* every pixel (x,y) in the window w *do*

$$W_{spatial} \rightarrow \exp\left(\frac{\|x_i - x_j\|^2 + \|y_i - y_j\|^2}{2\sigma_s^2}\right)$$

$$W_{spectral} \rightarrow \exp\left(\frac{\|I(x_i, y_i) - I(x_j, y_j)\|^2}{2\sigma_r^2}\right)$$

**Compute** $\sigma_h$ for class c

$$W_{nDSM} \rightarrow \exp\left(\frac{\|nDSM(x_i, y_i, t_m) - nDSM(x_j, y_j, t_n)\|^2}{2\sigma_h^2}\right)$$

W3D = $W_{Spatial} * W_{Spectral} * W_{nDSM}$

**Update** the probability distribution map
$$P_c^k(x_i, y_i, t_n) = \frac{1}{N*T} \sum \sum_{j \in N, n \in T} W_{3D}(x_j, y_j, t_n) * P_c^{k-1}(x_j, y_j, tn)$$

      **End For**

      **Check** convergence
$$\frac{P_c^k(x_i, y_i, t_n) - P_c^{k-1}(x_i, y_i, t_n)}{P_c^k(x_i, y_i, t_n)} * 100\% \rightarrow \begin{cases} \leq 5\% & Stop \\ > 5\% & Continue \end{cases}$$

   **End While**
*End For*
**Compute** overall accuracy

---

*Experimental Results and analysis*

The proposed filter was applied to three datasets that include an open area, residential area, and school area. The input data include multi-source and multitemporal very high-resolution images and DSMs; the probability maps were created for 6 types of classes: buildings, long term or temporary lodges, trees, grass, ground, and roads (see Figure 7(a) for more details about the input data). Figure 7(b) shows a sample of the filtering results, we can see that the initial classification of the building (circled with an ellipse) is mostly incorrectly classified to long-term lodge, however, it keeps improving as the filtering proceeds through the iterations. The overall accuracy was reported, and it indicates that the overall enhancement in the accuracy is about ~2-6% (see Table 3). We can also notice that dense areas as the residential area have the lowest accuracy range (around 85%), while the rest of the study areas had accuracy improvement in the 90% range. It indicates that the filtering algorithm is dependent on the degree of density and complexity in the scene, where objects are hard to distinguish in condensed areas due to mixed pixel and spectral similarity of different objects.

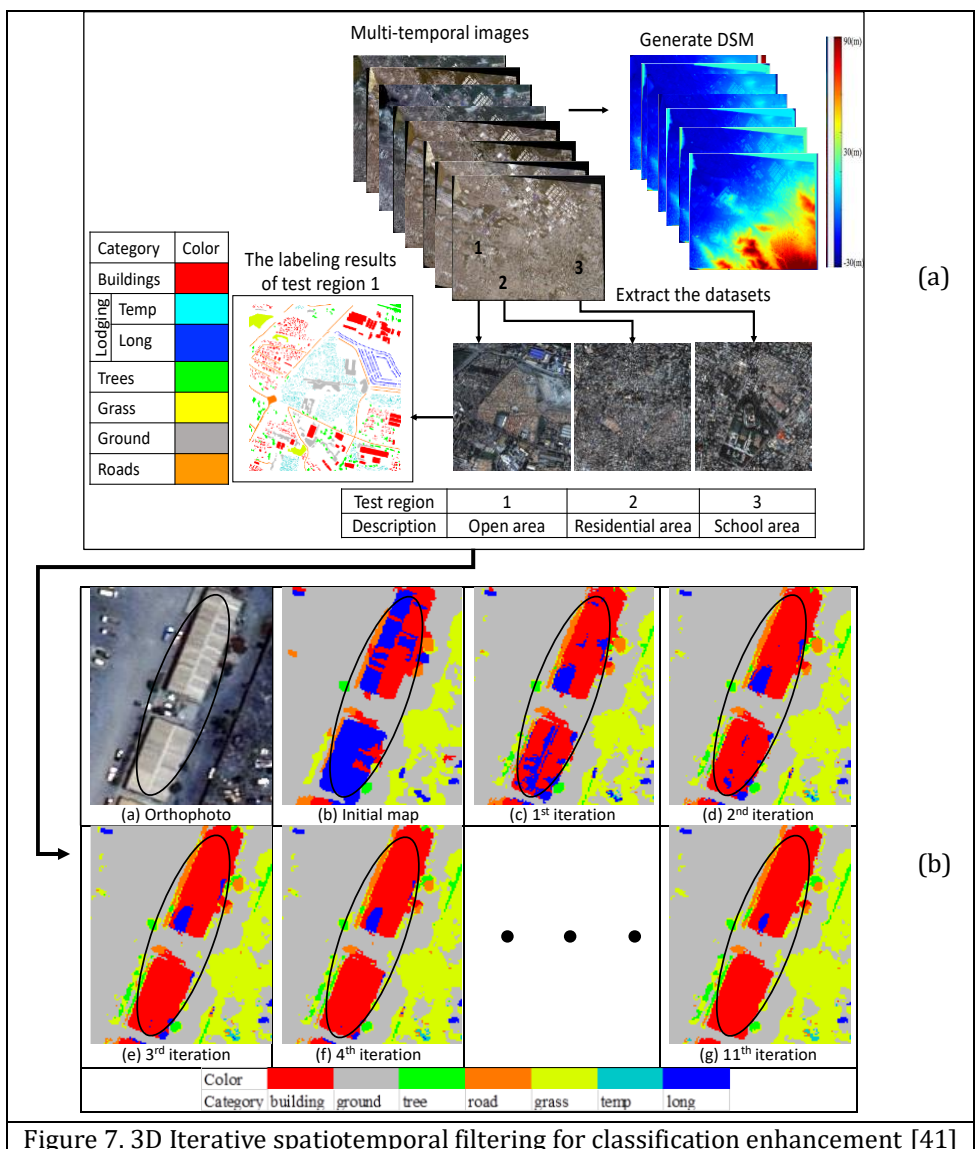

Figure 7. 3D Iterative spatiotemporal filtering for classification enhancement [41]

| Date | Test region 1 | | | Test region 2 | | | Test region 3 | | |
|---|---|---|---|---|---|---|---|---|---|
| | Before | After | Δ | Before | After | Δ | Before | After | Δ |
| 2007 | 91.04% | 95.21% | +4.17% | 83.47% | 88.14% | +4.67% | 91.12% | 95.85% | +4.73% |
| 2010/1 | 93.21% | 96.45% | +3.24% | 81.50% | 85.67% | +4.17% | 93.06% | 96.82% | +3.76% |
| 2010/6 | 91.93% | 96.26% | +4.33% | 83.52% | 89.79% | **+6.27%** | 88.82% | 94.87% | **+6.05%** |
| 2010/12 | 89.08% | 95.57% | **+6.49%** | 80.81% | 87.59% | **+6.78%** | 88.58% | 94.86% | **+6.28%** |
| 2012/3 | 92.19% | 95.92% | +3.73% | 81.43% | 86.92% | +5.49% | 91.44% | 97.08% | +5.64% |
| 2013/9 | 90.40% | 96.56% | **+6.16%** | 81.03% | 87.29% | **+6.26%** | 94.99% | 97.54% | +2.55% |
| 2014/7 | 95.11% | 97.27% | +2.16% | 82.19% | 88.90% | +6.17% | 90.39% | 96.58% | +6.19% |
| 2015 | 92.74% | 96.35% | +3.61% | 83.22% | 85.69% | +2.47% | 94.61% | 97.19% | +2.58% |
| Average | 92.09% | 96.20% | 4.24% | 82.15% | 87.50% | 5.29% | 91.63% | 96.35% | 4.72% |

Table 3. The overall accuracy for classification results using the method in [41]

### 4.3. Spatiotemporal fusion of 3D depth maps

*Background and objective*

Obtaining high-quality depth images (also known as depth maps) is essential for remote sensing applications that process 3D geometric information like 3D reconstruction. MVS algorithms are widely used approaches to obtain depth images (see Section 2.1.2.), however, depth maps generated using MVS often contain noise, outliers, and incomplete representation of depth like having missing data, holes, or fuzzy edges and boundaries. A common approach to recover the depth map is by fusing several depth maps through probabilistic or deterministic methods. However, most fusion techniques in image processing focus on the fusion of depth images from Kinect or video scenes, which cannot be directly applied on depth generated from satellite images due to the nature of images. The difference between depth generated from satellite sensors and Kinect or video cameras include:

1) Images captured indoor using Kinect or video cameras have less noise since they are not exposed to external environmental influences like atmospheric effects.
2) Kinect or video cameras generate a large volume of images, which can improve dense matching, while the number of satellite images is limited due to the temporal resolution of the satellite sensor.
3) The depth from satellite images is highly sensitive to the constant changes in the environment and the spatial characteristics of the earth surface like the repeated patterns, complexity, sparsity, and density of objects in the scene, which can obstruct or create mismatching errors in the dense image matching process.

Most depth fusion algorithms for geospatial data focus on median filtering (see Section 4.3.), but it still needs some improvement in terms of robustness and adaptivity to the scene content. To address the aforementioned problems, [90] proposed an adaptive and semantic-guided spatiotemporal filtering algorithm to generate a single depth map with high precision. The adaptivity is implemented to address the issue of varied uncertainty for objects of different classes.

*Theory*

The adaptive and semantic-guided spatiotemporal filter is a pixel-based fusion method, where the depth of the fused pixel is inferred using multitemporal depths and prior knowledge about the pixel class and uncertainty. A reference orthophoto is classified using a rule-based classification approach that uses normalized DSM (nDSM) with indices such as normalized difference vegetation index (NDVI). The uncertainty is then measured for all four classes (trees, grass, buildings, and ground and roads) using the standard deviation. The uncertainty is measured spatially using the classification map and also across the temporal images. The adaptive and semantic-guided spatiotemporal filter is intended to enhance the median filter, thus it uses height $h(i,j,t)_{med}$ as the base to the fused pixel, where the general form of the filter is expressed as

$$DSM_f(i,j) = \frac{1}{W_T} * \sum_{i=1}^{Width} \sum_{j=1}^{Height} W_r * W_s * W_h * h(i,j,t)_{med} \qquad (6)$$

Where $DSM_f$ is the fused pixel, $i, j$ are the pixels coordinates, $h_{med}$ is the median height value from the temporal DSMs, and the spectral, spatial, and temporal height weights are expressed as $W_r, W_s$ and $W_h$ respectively. The $W_r$ and $W_s$ are described in Eq. (3) and Eq. (4) that measure the spectral and spatial components from the orthophoto. The $W_h$ is a measure of similarity for the height data across temporal images, and it can be computed using the following formula

$$W_h(i,j) = exp^{\frac{-||hmed - h(i,j,t)||^2}{2\sigma_h^2}} \qquad (7)$$

Where $\sigma_h$ is the adaptive height bandwidth which varies based on the class of pixel as follows:

$$\sigma_h = \begin{cases} \sigma_{Building} \rightarrow if\ pixel\ (i,j)\ is\ building \\ \sigma_{Ground/road} \rightarrow if\ pixel\ (i,j)\ is\ ground/road \\ \sigma_{tree} \rightarrow if\ pixel\ (i,j)\ is\ tree \\ \sigma_{grass} \rightarrow if\ pixel\ (i,j)\ is\ grass \\ \sigma_{water} \rightarrow if\ pixel\ (i,j)\ is\ water \end{cases} \qquad (8)$$

*Experimental Results and analysis*

The method in [90] experimented on three datasets with varying complexities. The satellite images are taken from the World-View III sensor, and depth is generated using MVS algorithm on every image pair using RSP (RPC Stereo Processor) software developed by [95] and semi-global matching (SGM) algorithm [42]. Figure 8 describes the procedures followed by the fusion algorithm, in addition to the visual results where it shows that noise and missing elevation points were recovered in the fused image. The validation of three experiments shows that this fusion technique can achieve up to a 2% increase in the overall accuracy of the depth map.

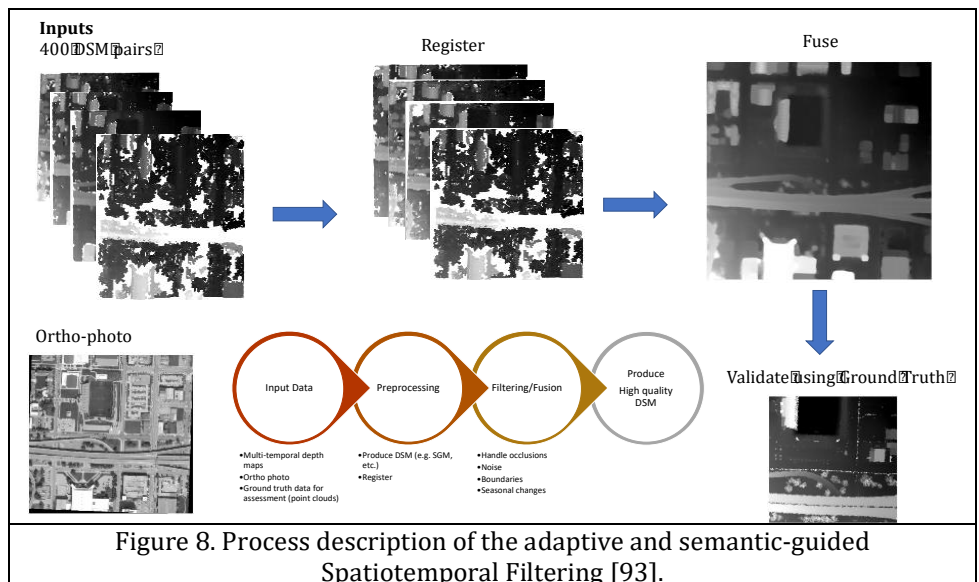

Figure 8. Process description of the adaptive and semantic-guided Spatiotemporal Filtering [93].

## 5. Conclusion(s)

Spatiotemporal fusion is one of the powerful techniques to enhance the quality of remote sensing data, hence, the performance of its applications. Recently, it has been drawing great attention in many fields, due to its capability to analyze and relate the space-time interaction on the ground that can lead to promising results in terms of stability, precision, and accuracy. The redundant temporal information is useful to develop a time-invariant fusion algorithm that leads to the same inference from the multitemporal geospatial data regardless of the noise and changes that occur occasionally due to natural (e.g., metrology, ecology, and phenology) or instrumental (e.g., sensor conditions) causes. Therefore, incorporating spatiotemporal analysis in any of the three levels of fusion can boost their performance, where it can be flexible to handle data from multiple sources, types, and times. Despite the effectiveness of spatiotemporal fusion, there are still some issues that may affect the precision and accuracy of the final output. These considerations must be taken into account while designing the spatiotemporal fusion algorithm. For example, spatiotemporal analysis for per-pixel operations is highly sensitive to mixed pixels especially for coarse resolution images where one pixel may contain the spectral information of more than one object. The accuracy of the spatiotemporal fusion can also be sensitive to the complexity of the scene, where in densely congested areas such as cities the accuracy may be less than open areas or sub-urban areas (as mentioned in the examples in Section 4.). This is due to the increase in the heterogeneity of the images in these dense areas. This issue can be solved using adaptive spatiotemporal fusion algorithms, which is a not widely investigated area of study in current practices. Feature and decision levels of fusion can partially solve this problem by learning from patches of features or classified images, but their accuracy will also be under the influence of the feature extraction algorithm or the algorithm to derive the initial output. For instance, mismatching features can result in fusing unrelated features or data points, thus produce inaccurate coefficients for the feature-level fusion model. Another observation is the lack of studies that relates the number of temporal images and the fusion output accuracy, which is useful to decide the optimal number of input images for fusion. Additionally, it is rarely seen that the integrated images are picked before fusion, where assessing and choosing good images can lead to better results. Spatiotemporal fusion algorithms are either local or global approaches, the local algorithms are simple and forward like pixel-level fusion or local filtering like the methods in [19, 22], while global methods tend to perform extensive operations for optimization purposes like in [25].  In future works, we aim to explore how these explicitly modeled spatiotemporal fusion algorithms can be enhanced by the power of more complex and inherent models such as deep learning based models to drive more important remote sensing applications.

## 6. Acknowledgments

The authors would like to express their gratitude for Planet Co. for providing us with the data, sustainable institute at the Ohio state university, Office of Naval Research (Award No. N000141712928) for partial support of the research, and the Johns Hopkins University Applied Physics Laboratory and IARPA and the IEEE GRSS Image Analysis and Data Fusion Technical Committee for making the benchmark satellite images available.